\pdfoutput=1
\documentclass[11pt]{article}

\usepackage{EMNLP2023}

\usepackage{times}
\usepackage{latexsym}
\usepackage{hyperref}
\usepackage{makecell}

\usepackage[T1]{fontenc}
\usepackage[utf8]{inputenc}

\usepackage{microtype}

\usepackage{multirow}

\usepackage{inconsolata}
\usepackage{tipa}
\usepackage{graphics}
\usepackage{graphicx}
\usepackage{wrapfig}
\usepackage{float}

\definecolor{NLPRed}{HTML}{FF0000}
\definecolor{NLPBlue}{HTML}{0000FF}

\newcolumntype{P}[1]{>{\centering\arraybackslash}p{#1}}

\title{\textit{Mergen}: The First Manchu-Korean Machine Translation Model\newline Trained on Augmented Data }

\author{Jean Seo, \ Sungjoo Byun, \ Minha Kang,  Sangah Lee  \\
         Seoul National University \\
         \texttt{\{seemdog, byunsj, alsgk1123, sanalee\}@snu.ac.kr}}

\begin{document}
\maketitle
\begin{abstract}

The Manchu language, with its roots in the historical Manchurian region of Northeast China, is now facing a critical threat of extinction, as there are very few speakers left. In our efforts to safeguard the Manchu language, we introduce \textit{Mergen}, the first-ever attempt at a Manchu-Korean Machine Translation (MT) model. To develop this model, we utilize valuable resources such as the M{\v a}nw{\' e}n L{\v a}od{\` a}ng(a historical book) and a Manchu-Korean dictionary. Due to the scarcity of a Manchu-Korean parallel dataset, we expand our data by employing word replacement guided by GloVe embeddings, trained on both monolingual and parallel texts. Our approach is built around an encoder-decoder neural machine translation model, incorporating a bi-directional Gated Recurrent Unit (GRU) layer. The experiments have yielded promising results, showcasing a significant enhancement in Manchu-Korean translation, with a remarkable 20-30 point increase in the BLEU score.

\end{abstract}

\section{Introduction}

Efforts to conserve and revive endangered languages have surged, with modern advancements in Natural Language Processing (NLP) playing a pivotal role. \citet{Cherokee} introduce ChrEn, a Cherokee-English parallel dataset, and examine methodologies like Statistical Machine Translation (SMT) and Neural Machine Translation (NMT).  \citet{Cherokee} aid the conservation of Cherokee, a critically endangered Native American dialect. On a similar note, \citet{Deciphering} present a decipherment model for lost languages that addresses challenges posed by non-segmented scripts and undetermined proximate languages, leveraging linguistic constraints and the International Phonetic Alphabet (IPA) for phonological patterns.

Manchu language, originated from the historical Manchurian region in Northeast China, stands as a highly endangered Tungusic language of East Asia \citep{manchu_endangered}. There are merely few Manchu speakers left nowadays, leading Manchu to be labeled `nearly extinct' by UNESCO \citep{unesco}. The Manchu spell checker \citep{speller} and the Manchu corpus with morphological annotations \citep{ChoiEtAl2023a, ChoiEtAl2023b} are the only prior approaches to embrace Manchu in the field of NLP. We introduce \textit{Mergen}, the first Manchu-Korean machine translation model, which marks the pioneering effort to apply MT to the Manchu language.

We employ two sets of parallel corpora for machine translation from Manchu to Korean, as detailed in \citet{KimEtAl2019}. Initially, we train an adapted version of the NMT model \citep{NMT}. Assuming the unexpectedly low performance is due to the scarcity of Manchu-Korean data, we augment the size of parallel data several fold utilizing GloVe \citep{pennington-etal-2014-glove}. Our findings suggest that this data augmentation methodology substantially enhances translation quality. 

Despite the constrained availability of resources, our goal is to enhance Manchu-Korean machine translation performance. To symbolize our commitment to the field of Manchu NLP, we christen our model \textit{Mergen}, denoting a sage or a wise individual in the Manchu lexicon. Our translation approach, which employs a data augmentation technique, not only seeks to improve Manchu-Korean translation performance but also aims to eventually serve as a potential model for addressing NLP challenges in other extremely low-resource scenarios as addressed in \citet{low}.

\section{Related Work}
\label{sec:2-Related Work}
\subsection{Low-Resource Machine Translation}
\label{sec:2-1-LRMT}
MT necessitates parallel data of source and target languages to be trained effectively. However, the majority of language pairs face a scarcity of resources. As a result, there has been various research endeavors aimed at developing translation models in low-resource scenarios. Extended language models such as XLM-RoBERTa \citep{xlm-r}, mBART \citep{mBART50}, multilingual BERT (mBERT) \citep{mBERT}, and mT5 \citep{mT5} are trained on diverse languages. Yet, most of these multilingual language models tend not to incorporate endangered languages. This leads to an increasing disparity in NLP resources, where less-resourced languages are further marginalized. Numerous strategies have been attempted in low-resource machine translation. \citet{monolingual} and \citet{monolingual2} employ monolingual data in low-resource NMT. Additionally, utilization of pre-trained word embeddings \citep{pretrained_word_embeddings} and application of transfer learning with pretrained language models like XLM \citep{xlm} and mBART \citep{mbart} have been employed. Furthermore, \citet{zeroshot} enhance the zero-shot translation capability of low-resource languages.

\begin{figure}[t]
    \centerline{\includegraphics[width=\columnwidth]{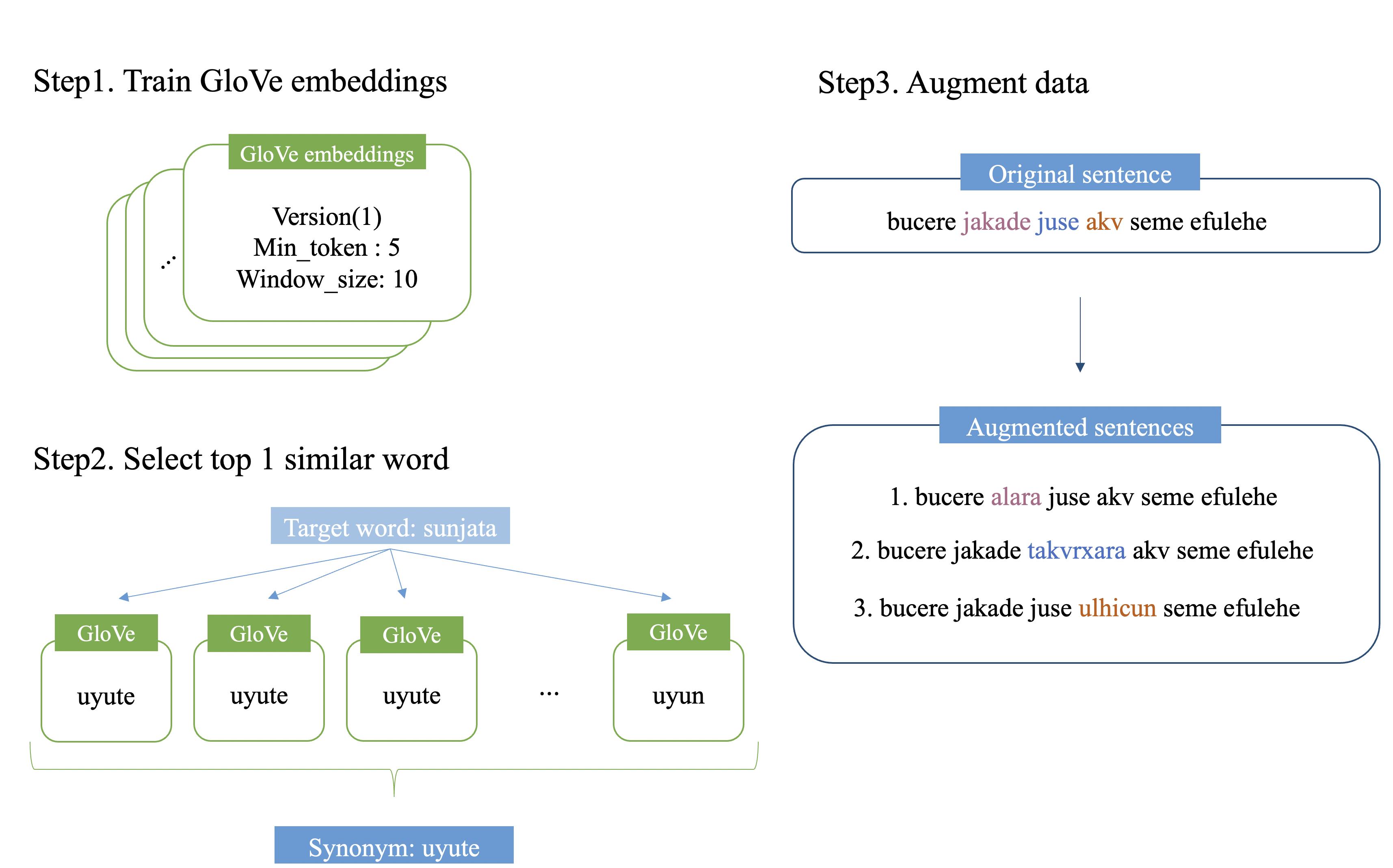}}
    \caption{Our data augmentation methodology. First, we train ten versions of GloVe embedding models, varying in the minimum token length of source data and window size. Then, the presumable synonym for the target word is selected via comparing the frequency of outputs from each model. Finally, we augment data through replacing original words with synonyms if possible. The pair of original and substituted words are in the same color.}
    \label{figure_1} 
\end{figure}

\subsection{Typological Similarities between Manchu and Korean}
\label{sec:2-2-Typology}
There are several typological motivations for translating Manchu to Korean using a Machine Translation model. The genetic affinity between Manchu and Korean is not proven, but it is well-known that Manchu has a similar structure to that of Korean. The word order of Manchu and Korean mostly coincide, including the order of ‘noun-particle,’ ‘modifier-modified,’ and ‘object-verb,’ etc. \citep{ParkSangChul2018}. Substitutes in Korean, \textit{kes}, and Manchu, \textit{-ngge}, have analogous grammatical functions and positions \citep{ChoiDongGuen2009}. The two languages both show factivity alternation by using the attitude verb ‘to know’ \citep{LeeJungMin2019} and have parallel subordinated clause structures \citep{MalchukovAndCzerwinski2020}. These typological similarities between Manchu and Korean arouse interest in understanding and linguistically translating each other. In fact, studies of the Manchu language are active in Korea \citep{KoDongHo2023}.

\begin{table}[b]
\resizebox{\columnwidth}{!}{%
\begin{tabular}{cc}
\Xhline{3\arrayrulewidth}
\textbf{Monolingual data} & \textbf{Number of sentences} \\
\hline
M{\v a}nw{\' e}n L{\v a}od{\` a}ng--Taizong & 2,220 \\
Ilan gurun i bithe & 41,904 \\
Gin ping mei bithe & 21,376 \\
Y{\` u}zh{\` i} Qīngw{\' e}nji{\` a}n & 11,954 \\
Y{\` u}zh{\` i} Z{\= e}ngd{\` i}ng Qīngw{\' e}nji{\` a}n & 18,420 \\
\hline
\textbf{Parallel data (Man-Kor)} & \\
\hline
M{\v a}nw{\' e}n L{\v a}od{\` a}ng--Taizu & 22,578 \\
Manchu-Korean Dictionary & 40,583 \\
\Xhline{3\arrayrulewidth}
\end{tabular}%
}
\caption{The size of each material}
\label{tab:size}
\end{table}

\section{Data}
\subsection{Materials}
The Manchu corpora used in this study comprise all of the digitized textual data available and can be categorized as either parallel or monolingual. The parallel corpora are \textit{M{\v a}nw{\' e}n L{\v a}od{\` a}ng} (1774-1778) and the Manchu-Korean dictionary. These corpora consist of Manchu texts and their corresponding translations in Korean. We only utilize a section of the \textit{M{\v a}nw{\' e}n L{\v a}od{\` a}ng} and its translations from \citet{KimEtAl2019}, which details the history of Nurhaci, the Emperor Taizu of Qing dynasty. Additionally, we refer to the dictionary from \citet{LeeHoon2017} and select sentences with a minimum of three words.

The monolingual texts of Manchu include the remaining part of \textit{M{\v a}nw{\' e}n L{\v a}od{\` a}ng}, Manchu-Manchu dictionaries, and several pieces of literature. The part of \textit{M{\v a}nw{\' e}n L{\v a}od{\` a}ng} left over is the chronicle of Hong Taiji, the Emperor Taizong of Qing. The Manchu-Manchu dictionaries we use are \textit{Y{\` u}zh{\` i} Qīngw{\' e}nji{\` a}n} (1708) and \textit{Y{\` u}zh{\` i} Z{\= e}ngd{\` i}ng Qīngw{\' e}nji{\` a}n} (c.1771).

The other data is composed of novels, \textit{Ilan gurun i bithe} (c.1723-1735) and \textit{Gin ping mei bithe} (1708). \textit{Ilan gurun i bithe} is the translated version of \textit{The Romance of the Three Kingdoms}. \textit{Gin ping mei bithe} is translated from the Chinese naturalistic novel, \textit{The Plum in the Golden Vase}. The size description of each data can be found in Table~\ref{tab:size}.

\subsection{Romanization of Manchu script and Hangul}
To create a more sufficient translation model, the script of each language should be unified in one writing system. That is, both the source and target language should undergo transliteration to the Latin alphabet, so-called ‘romanization’. For the romanization of Manchu, we apply Abkai Latin transliteration. The Abkai romanization suggested by \citet{An1993} is a Pinyin-based writing system. We also use the system of \citet{SeongBaegin1977} for the special characters in the Manchu script. Transliteration of Manchu to the Latin alphabet is reversible except for a couple of letters. For the Latin transliteration of Korean, we employ Yale romanization system \citep{Martin1992} and develop the corresponding Python library\footnote{https://github.com/SeHaan/YaleKorean}. See Appendix ~\ref{sec:appendix} for examples.

\subsection{Data Augmentation}
The lack of available Manchu linguistic data poses challenges not only for the pre-training of transformer-based models but also for the training of simpler and more lightweight models, such as encoder-decoder models. Inspired by TinyBERT \citep{jiao2020tinybert}, we adopt a novel data augmentation approach. While the data augmentation method in TinyBERT \citep{jiao2020tinybert} combines both BERT \citep{devlin2019bert} and GloVe \citep{pennington-etal-2014-glove}, we exclusively employ GloVe embeddings. This decision stems from the absence of a pre-trained BERT model tailored to Manchu and the significant difficulty of pre-training a BERT model from scratch due to the limited amount of available textual data.

Our methodology involves training GloVe embedding models with two different versions of the dataset: (1) a dataset comprising sentences with at least 3 words, and (2) a dataset comprising sentences with at least 5 words. The dataset includes both monolingual and parallel text data. Various window sizes, specifically 1, 3, 5, 7, and 10, are used during the training process, resulting in a total of 10 distinct variations of GloVe embeddings. 

For each word in the training dataset, we gather the most similar word predicted by each individual GloVe embedding. Amongst the list of 10 words generated from these separate models, the word with the highest frequency is considered the most suitable synonym for the target word. Following this, we substitute a single word in each sentence from parallel text data with the identified synonym. The augmentation steps are described in Figure \ref{figure_1}. This procedure leads to the creation of two augmented versions of the original dataset: full augmentation and half augmentation. The first version involves replacing every word possible in each sentence with its corresponding synonym, significantly expanding the dataset size relative to the average sentence length. The second version is generated by replacing half of the words in each sentence with their respective synonyms, resulting in a dataset expansion about half the size of the first method. Additional details regarding the original and augmented dataset are available in Table~\ref{tab:augmented data}.

\begin{table}[ht]
\resizebox{\columnwidth}{!}{%
\begin{tabular}{cP{3cm}c}
\Xhline{3\arrayrulewidth}
\textbf{augmentation} & \textbf{M{\v a}nw{\' e}n L{\v a}od{\` a}ng\newline--Taizu (train)}
& \textbf{Man-Kor Dict}  \\ 
\hline
\textbf{Before augmentation}   & 20,320                          & 40,583                 \\
\textbf{Full augmentation}     & 179,843                         & 154,404                \\
\textbf{Half augmentation}     & 99,506                          & 100,694     
 \\
\Xhline{3\arrayrulewidth}
\end{tabular}%
}
 \caption{The number of sentences of parallel text data before and after augmentation}
 \label{tab:augmented data}
\end{table}

\section{Experiments}
\subsection{Task Details}
In the experiment, we merge \textit{M{\v a}nw{\' e}n L{\v a}od{\` a}ng} with Manchu-Korean dictionary and shuffle them together. The combined dataset is then divided into training, validation, and testing subsets. These subsets are split in an 8:1:1 ratio. In the augmentation process, we first shuffle and then augment the data to even out the word distributions, finally splitting into subsets.

\subsection{Model}
We adopt the sequence-to-sequence (seq2seq) framework, a deep learning approach designed to transform one sequence into another. Our model is based on the encoder-decoder structure of the NMT \citep{NMT}, implemented with bi-directional Gated Recurrent Unit (GRU) layer \citep{baseline}. We incorporate two techniques to enhance the performance: packed padded sequences and masking. Packed padded sequences ensure that the RNN processes only the genuine elements of the input sentence, excluding the padded ones. Masking directs the model to deliberately overlook specific components, like attention weights assigned to padded sections. 

\subsection{Results and Discussions}
% We use BLEU score \citep{bleu} to evaluate the performance of the machine translation.
% \textred{\textit{to be continued...}}

% \begin{table}
% \scriptsize
% \centering \resizebox{\columnwidth}{!}{
% \begin{tabular}{p{3cm}cc}
% \hline
% \textbf{Model} & \textbf{BLEU} & \textbf{PPL}\\
% \hline
% encoder-decoder \newline
% with GRU layer & 00 & 00 \\\hline
% \end{tabular}}
% \caption{Model Performance: Evaluating BLEU score and perplexity on the test dataset}
% \label{tab:bleu}
% \end{table}

\begin{table}[tbp]
\centering
\resizebox{\columnwidth}{!}{%
\begin{tabular}{lccc}
\Xhline{3\arrayrulewidth}
\textbf{Train} & \textbf{Test} & \textbf{BLEU} & \textbf{PPL} \\
\Xhline{3\arrayrulewidth}
\multicolumn{4}{l}{\textbf{Before augmentation (No augmentation)}} \\
\hline
M{\v a}nw{\' e}n L{\v a}od{\` a}ng & M{\v a}nw{\' e}n L{\v a}od{\` a}ng & 0.0 & 72.50 \\
Man-Kor Dict & Man-Kor Dict & 0.0 & 59.34 \\
\hline
\multirow{3}{*}{Combined} & M{\v a}nw{\' e}n L{\v a}od{\` a}ng & 0.0 & 61.83 \\
& Man-Kor Dict & 0.0 & 61.16 \\
& Combined & 0.0 & 69.62 \\
\Xhline{3\arrayrulewidth}
\multicolumn{4}{l}{\textbf{Half augmentation}} \\
\hline
M{\v a}nw{\' e}n L{\v a}od{\` a}ng & M{\v a}nw{\' e}n L{\v a}od{\` a}ng & 38.38 & 147.07 \\
Man-Kor Dict & Man-Kor Dict & 0.0 & 174.94 \\
\hline
\multirow{3}{*}{Combined} & M{\v a}nw{\' e}n L{\v a}od{\` a}ng & 36.05 & 192.95 \\
& Man-Kor Dict & \textbf{2.37} & 36.14 \\
& Combined & 27.59 & 29.22 \\
\Xhline{3\arrayrulewidth}
\multicolumn{4}{l}{\textbf{Full augmentation}} \\
\hline
M{\v a}nw{\' e}n L{\v a}od{\` a}ng & M{\v a}nw{\' e}n L{\v a}od{\` a}ng & \textbf{38.95} & 1549.40 \\
Man-Kor Dict & Man-Kor Dict & 0.0 & 158.25 \\
\hline 
\multirow{3}{*}{Combined} & M{\v a}nw{\' e}n L{\v a}od{\` a}ng & \textbf{37.17} & 447.59 \\
& Man-Kor Dict & 2.26 & 46.54 \\
& Combined & \textbf{28.00} & 41.97 \\ \hline
\Xhline{3\arrayrulewidth}
\end{tabular}
}
\caption{Manchu-Korean Translation Performance}
\label{tab:results}
\end{table}

We perform machine translation and evaluate the performance on all the available combinations of parallel corpora: \textit{M{\v a}nw{\' e}n L{\v a}od{\` a}ng}, Manchu-Korean dictionary, and the combined dataset. In particular, we augment the training sets of each corpus to alleviate the data scarcity problem. \autoref{tab:results} shows the performance of our Manchu-Korean translation models, with BLEU score \citep{bleu} and Perplexity (PPL) as the metrices. We train each model for 5 epochs and report the one with the best performance.

The first block of \autoref{tab:results} shows the translation performance based on the original Manchu-Korean parallel corpora. All the experiments here show BLEU scores of 0.0, which represent that none of the test sentences are accurately translated. Most of the predicted translations include the special symbol `<UNK>' instead of proper Korean tokens, possibly due to the small dataset and vocabulary size.

The second block shows the experiment results from the augmented version of the parallel corpora, where up to 50\% of the tokens in each sentence are replaced for data augmentation. The third block displays experiments on another augmented version where all tokens with substitutes are replaced. The augmentation procedure increases the size of the training set, resulting in a significant rise in the translation performance. BLEU scores exceed 38 on the \textit{M{\v a}nw{\' e}n L{\v a}od{\` a}ng} test set, and around 28 on the combined test set. The two versions of the augmented dataset show comparable performance, but replacing all the possible words in the corpus resulted in slightly higher BLEU scores.

Due to data augmentation, the vocabulary for each model is expanded; for example, the original \textit{M{\v a}nw{\' e}n L{\v a}od{\` a}ng} vocabulary includes 4,335 words, while the full-augmented dataset constructs an expanded vocabulary with 11,089 words. A larger vocabulary and training set may have helped the language model's representation and result in better translation performance. Additionally, most newly induced words are from the augmentation sources which include monolingual Manchu texts, different from our parallel corpora. This expansion of word diversity may have also affected the models' perplexity to increase when they predicted the next words in each sentence.

On the other hand, results on the Manchu-Korean dictionary are consistently very low, and this may have influenced the lower performance of the combined test set. We suppose that it is because the corpus is a dictionary, where each line is a unique word or phrase. The training set and the test set would have much fewer overlaps in their vocabularies, and this could cause a number of `<UNK>' generations in the model prediction.

\section{Conclusion}
In our exploration of the critically endangered Manchu language, we have made significant strides towards development of low-resource NLP through the development of the Manchu-Korean MT system, "Mergen." Our endeavor to train this model, despite the challenges posed by the scarcity of a Manchu-Korean parallel dataset, demonstrates the potential of an innovative data augmentation strategy. This attempt is also significant in that we have collected all the digitized Manchu text data. By leveraging resources such as "Mǎnwén Làodǎng" and a Manchu-Korean dictionary, and by adopting a word substitution techniqus guided by GloVe embeddings, we have not only built a functional MT system but have also considerably enhanced its accuracy, as evidenced by the increase in the BLEU score. Our encoder-decoder NMT model, equipped with a bi-directional GRU layer, has shown promising results, offering hope for the preservation and accessibility of the Manchu language to future generations. We anticipate that this research will serve as a foundation for further innovations in the realm of endangered language preservation.

\section*{Limitations}
The main limitation of this study is the scarcity of resources. Numerous Manchu literatures exist in East Asia \citep{Vovin2024}, including China \citep{elliott2001manchu}, Korea \citep{KoAndYou2012}, and Mongolia \citep{Choi2014}. However, most of them lack an electronic version. The only publicly available Manchu language database is the Manchu Dictionary and Literature DB, created by Seoul National University and supported by the National Research Foundation of Korea.\footnote{NFR-2012S1A5B4A01035397, available at \url{http://ffr.krm.or.kr/base/td037/intro_db.html}} Furthermore, the majority of these resources have not been translated into Korean. To address this gap, we intend to provide supplementary parallel texts translated into Korean for further study. In addition, we plan to implement a cutting-edge method of Transformer-based language model including Manchu language. Knowledge Distillation could be a way for modeling endangered languages, training a small student model based on those languages and improving it with a teacher model based on high-resource languages \citep{heffernan}.

\section*{Ethics Statement}
The Manchu language, classified as critically endangered, remains underrepresented due to its scarce resources. As such, it has yet to be incorporated into any multilingual language models. This study pioneers Manchu translation efforts, an endeavor previously uncharted. Our primary research objective as NLP practitioners is to prevent the extinction of Manchu language and ensure its preservation. We have no intention of commercializing the translation model. Instead, by making the model publicly available, we aim to facilitate and encourage as many individuals as possible to learn Manchu using our translator. We are committed to continuous collaboration with Manchu language researchers. We endeavor to enhance the performance of our translator and regularly update it with new Manchu data to ensure its accuracy.

\bibliography{anthology,main}
\bibliographystyle{acl_natbib}

\appendix

\newpage
\section{Example Appendix}
\label{sec:appendix}

\begin{wrapfigure}{i}{\columnwidth}
    \centerline{\includegraphics[width=\columnwidth]{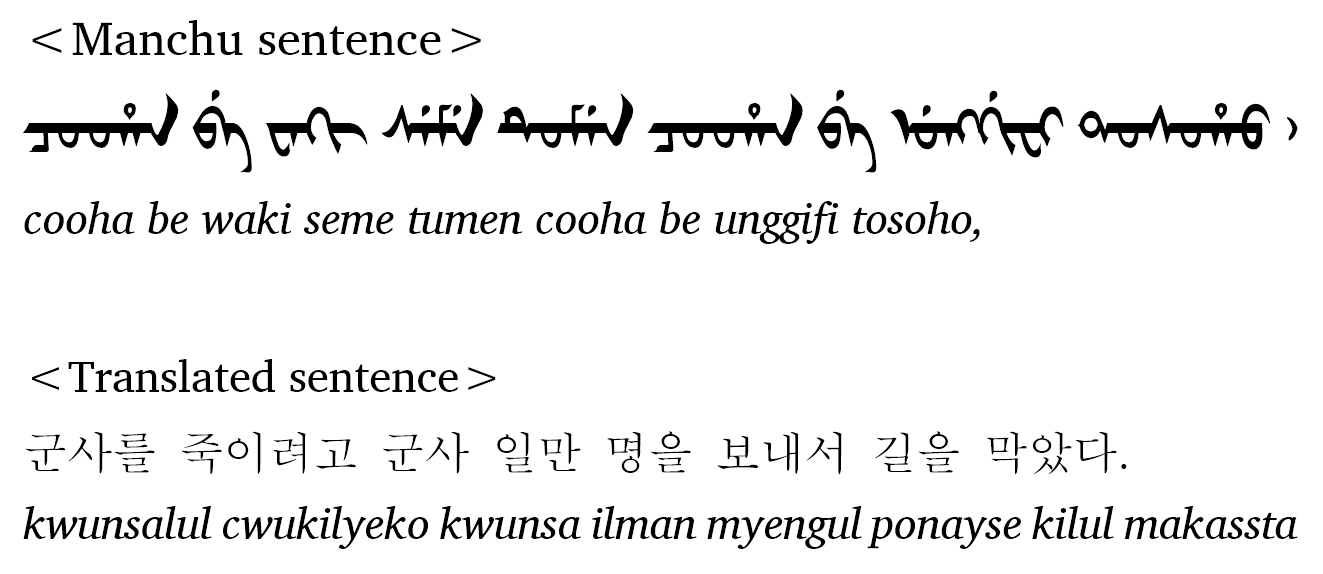}}
    \caption{Example of Romanizations of Manchu text and Korean text}
\end{wrapfigure}

\end{document}